\documentclass[10pt,journal]{IEEEtran}

\usepackage[table]{xcolor}
\usepackage{xspace}
\usepackage{bm}
\usepackage{amsfonts}
\newcommand{\zs}{\texttt{zoNNscan}\xspace}
\usepackage{graphicx}

\usepackage{flushend}

\newtheorem{definition}{Definition}
\usepackage{algorithm}
\usepackage{algorithmic}

\begin{document}

\title{\zs: a boundary-entropy index\\ for zone inspection of neural models}


\author{\IEEEauthorblockN{Adel Jaouen\thanks{A. Jaouen and E. Le Merrer are  with  Technicolor, Cesson-S\'evign\'e, 35576, France.  (e-mail: erwan.lemerrer@technicolor.com)} and}
  \and
  \IEEEauthorblockN{Erwan Le Merrer \footnotemark}
}

\maketitle

\begin{abstract}
The training of deep neural network classifiers results in decision boundaries which geometry is still not well understood. This is in direct relation with classification problems such as so called adversarial examples. 
We introduce \zs, an index that is intended to inform on the boundary uncertainty (in terms of the presence of other classes) around one given input datapoint. It is based on confidence entropy, and is implemented through sampling in the multidimensional ball surrounding that input. 
We detail the \zs index, give an algorithm for approximating it, and finally illustrate its benefits on four applications, including two important problems for the adoption of deep networks in critical systems: adversarial examples and corner case inputs. We highlight that \zs exhibits significantly higher values than for standard inputs in those two problem classes. 
\end{abstract}


\section{Introduction}
\label{sec:introduction}

Measures are important tools in all computer science domains, as far as they allow to quantify and then analyze specific problems \cite{995823}, or to serve as a building block for designing better algorithms \cite{941248}.
In the blooming field of machine learning leveraging deep neural networks, open problems remain that are related to the emergence of \textit{decision boundaries} in classification tasks. Little is known about the geometrical properties of those \cite{DBLP:journals/corr/FawziMFS17}.
Yet, the relation between the position of those boundaries, with respect to classification of individual input datapoints at inference time, is getting more salient with the increasing amount critical applications that are using deep neural networks.
The case of self-driving cars is an illustration of a critical application, where corner-cases have been recently found in production models (leading to wrong decisions) \cite{deepxplore}; the input that causes the erroneous classification are depicted to be close from decision boundaries. Another example class of problems is known as adversarial examples \cite{Goodfellow:2015}, where inputs are slightly modified in order to cause the target model to misclassify them. The modified inputs are also close to the boundary of their original classes, as humans barely see the modification of those inputs. In addition, there are also recent works that are voluntarily tweaking the decision boundaries around given inputs, so that ownership information can be embedded into a target model \cite{watermarking}.

While measures are inherently in use with neural networks to evaluate the quality of the learning task over a given dataset, we find that there is a lack of an index that provides information on the neighboring of given inputs, with regards to the boundaries of other classes. 
Visual inspections of suspicious zones, by plotting in 2D a given decision boundary and the considered inputs is possible \cite{plotting}. In this paper, we propose an algorithm to output an index, for it helps to automatically identify problems at inference time.

The remaining of this paper is structured as follows. Section \ref{s:metric} introduces the \zs index, as well as a Monte Carlo approximation of it. The use of \zs is then demonstrated~
\footnote{The implementation of \zs is open-sourced on GitHub \cite{source}.} 
in Section \ref{s:experiments}, on four different scenarios. We review related works and conclude in Sections \ref{s:related} and \ref{s:conclusion}.
 \begin{figure}[t!]
\centering
\includegraphics[width=0.5\linewidth]{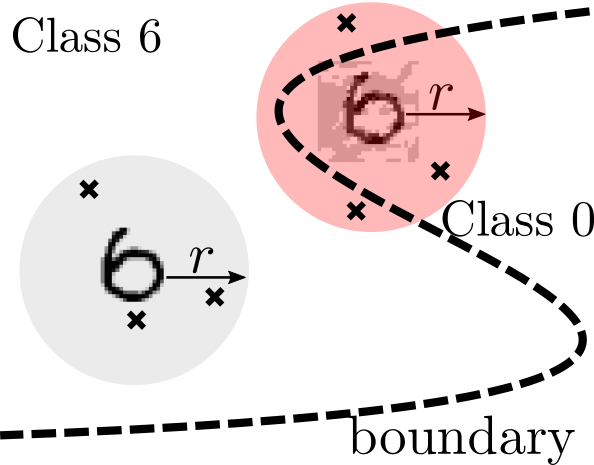}
\caption{An illustration of \zs runs around two inputs. The example on the left (a $6$ digit) is far from the model decision boundary, thus a \zs at radius $r$ returns a value close to $0$. The example on the right is an adversarial example (causing the 6 digit to be misclassified as a 0), that is by definition close to the decision boundary of its original class; \zs returns a higher value (up to 1), as some data samples in the ball fall into the original digit class.}
\label{fig:boundary}
\vspace{-0.5cm}
\end{figure}

\section{Zoning classification uncertainty}
\label{s:metric}

Classically, given an input and a trained model, the class predicted by the classifier is set to be the one with the higher score in the output vector \cite{insights}. These scores can thus be considered as membership probabilities to the model classes:
output values close to 1 are to be interpreted as a high confidence about the prediction, and conversely, values close to 0 show high probabilities of non membership to the corresponding class. These scores also provide the information about the \textit{uncertainty}  of the model for a given input : vector scores close to each other indicate an uncertainty between these classes, and equal scores characterize a decision boundary point \cite{554193}. Thereby, a maximum uncertainty output refers to an input that causes inference to return an uniform vector of probabilities $\frac{1}{C}$ for a classification task into $C$-classes.
Conversely, minimum uncertainty corresponds to an input that causes inference to return a vector of zeros, except for the predicted class for which a one is set.
\zs evaluates the uncertainty of predictions in a given input region.
This is illustrated on Figure \ref{fig:boundary}: the digit on the left is surrounded by no decision boundary in a ball of radius $r$ around it. The digit on the right is an adversarial example crafted from the previous digit; it is now classified as a 0, and a \zs in a ball of the same radius indicates the presence of boundary and then a possibly problematic input with higher \zs value.

\subsection{\zs: definition}
Given a discrete probability distribution $p = \{p_1, ..., p_C\} \in [0,1]^C$ with $\Sigma_{i=1}^{C}p_i = 1$ ($C$ designates the number of events), we remind the definition of the Shannon entropy ($b$ designates the base for the logarithm) :

$\begin{array}{clcl}
  H_b : &[0,1]^{C}  &\rightarrow  &\mathbb{R^{+}}\\
      &p          &\mapsto      &\Sigma_{i=1}^C(-p_i \log_b(p_i)).\\
\end{array}$ \\

The maximum of $H_b$ is reached in $p = \{ \frac{1}{C}, ..., \frac{1}{C} \}$ and is equal to $\log_b(C)$ and with the convention $0 \log_b(0) = 0$, the minimum of $H_b$ is 0. Subsequently, we use a $C$-based logarithm to make $H_C$ output values in range $[0,1]$.
A classifier $\mathcal{M}$ is considered as a function defined on the input space $[0,1]^d$ (on the premise that data are generally normalized for the training phase in such a way that all features are included in the range $[0,1]$), taking its values in the space $\mathbb{P}=\{p \in [0,1]^C~s.t.~\Sigma_{i=1}^{C}p_i = 1\}$. We introduce the composite function $\varphi = H_b\circ\mathcal{M}$ to model the indecision the network has on the input space. More specifically, we propose the expectation of $\varphi(U)$, $\mathbb{E}_{\mathbb{Z}}[\varphi(U)]$ (with $U$ a uniform random variable on $\mathbb{Z}$), on an input zone $\mathbb{Z}\in[0,1]^d$, to be an indicator on the uncertainty of the classifier on zone $\mathbb{Z}$.

\begin{definition}{Let the \zs index be, in zone $\mathbb{Z}$:}

$\begin{array}{clcl}
\scriptsize
   &\mathcal{B}([0,1]^d)  &\rightarrow  &[0,1]\\
      &\mathbb{Z}          &\mapsto      &\mathbb{E}_{\mathbb{Z}}[\varphi(U)] = \displaystyle \int_{\mathbb{Z}} \varphi(u)f_U(u) \, \mathrm{d}u,\\
\end{array}$ \\

where $\mathcal{B}(\mathbb{R}^d)$ refers to the $\mathbb{R}^d$ Borel set and $f_U$ the uniform density on $\mathbb{Z}$. 
\end{definition}

The minima of $\mathbb{E}_{\mathbb{Z}}[\varphi(U)]$ depicts observations in $\mathbb{Z}$ that were all returning one confidence of $1$, and $C-1$ confidences of $0$. Conversely, the maxima indicates full uncertainty, where each observation returned $\frac{1}{C}$ confidence values in the output vector.

\subsection{A Monte Carlo approximation of \zs}
In practice, as data are typically nowadays of high dimensionality  (e.g., in the order of millions of dimensions for image applications) and deep neural networks are computing complex non-linear functions, one cannot expect to compute the exact value of this expectation. 
We propose a Monte Carlo method to estimate this expectation on a space corresponding to the surrounding zone of a certain input.

For inspection around a given input $X\in[0,1]^d$, we consider a ball $\mathbf{B}$ of center $X$ and radius $r$, as zone $\mathbb{Z}$ introduced in previous subsection.
We perform a Monte Carlo sampling of $k$ inputs in a ball for the infinite-norm, corresponding to the hyper-cube of dimension $d$, around $X$ (as depicted on Figure \ref{fig:boundary}).
We are generating inputs applying random deformations\footnote{For more advanced sampling techniques in high dimensional spaces, please refer to e.g., \cite{dick_kuo_sloan_2013}.} $\epsilon_i$ on each components $X_i$ such as $max(-X_i,-r)\leq\epsilon_i\leq min(1-X_i,r)$. 
 
 For instance, given a normalized input $X$ and a positive radius $r$, Monte Carlo sampling is performed uniformly in the subspace of $\mathbb{R}^d$ defined as: $$\mathbb{Z} = \mathbf{B}_{\boldsymbol{\infty} }(X,r)\cap[0,1]^d.$$

The larger the number of samples $k$, the better the approximation of the index; this value has to be set considering the application and the acceptable computation/time constraints for inspection.


The \zs index, for inspection of the surrounding of input $X$ in a model $\mathcal{M}$, is presented on Algorithm \ref{algo}.

\begin{algorithm}[t!]
\begin{algorithmic}

\REQUIRE $\mathcal{M}, X, r, k$
\FOR{$i=0..k$}
\STATE $x' \gets$ Monte Carlo Sampling in $ \mathbf{B}_{\boldsymbol{\infty} }(X,r)\cap[0,1]^d$
\STATE $P_i \gets \mathcal{M}(x')$ \COMMENT{Inference}
\STATE $entropy_i \gets H_C(P_i)$ 
\ENDFOR
\RETURN $\frac{1}{k} \sum\limits_{i=1}^k entropy_i$

\end{algorithmic}
\caption{Monte Carlo approximation of \zs}
\label{algo}
\end{algorithm}

\section{Illustration use of \zs}
\label{s:experiments}

We first give an example of \zs use for the analysis of the surrounding regions of four inputs from the MNIST and CIFAR-10 datasets, in subsection \ref{ss:uncertainty}, using the multi-layer perceptron (noted MLP hereafter) proposed as an example in the 
Keras library \cite{keras}. We then give two illustrations of the capacity of \zs to identify problematic zones around inputs, in \ref{ss:adv} and \ref{ss:disagreement}. We conclude this section in \ref{ss:watermarking} by considering the impact of a novel watermarking technique on boundaries.
In the sequel, if not mentioned, Algorithm \ref{algo} is run with $k=10,000$ samples.

\begin{figure*}[t!]
\centering
\begin{minipage}{.49\linewidth}%
\hspace{-0.6cm}\includegraphics[width=1.07\linewidth]{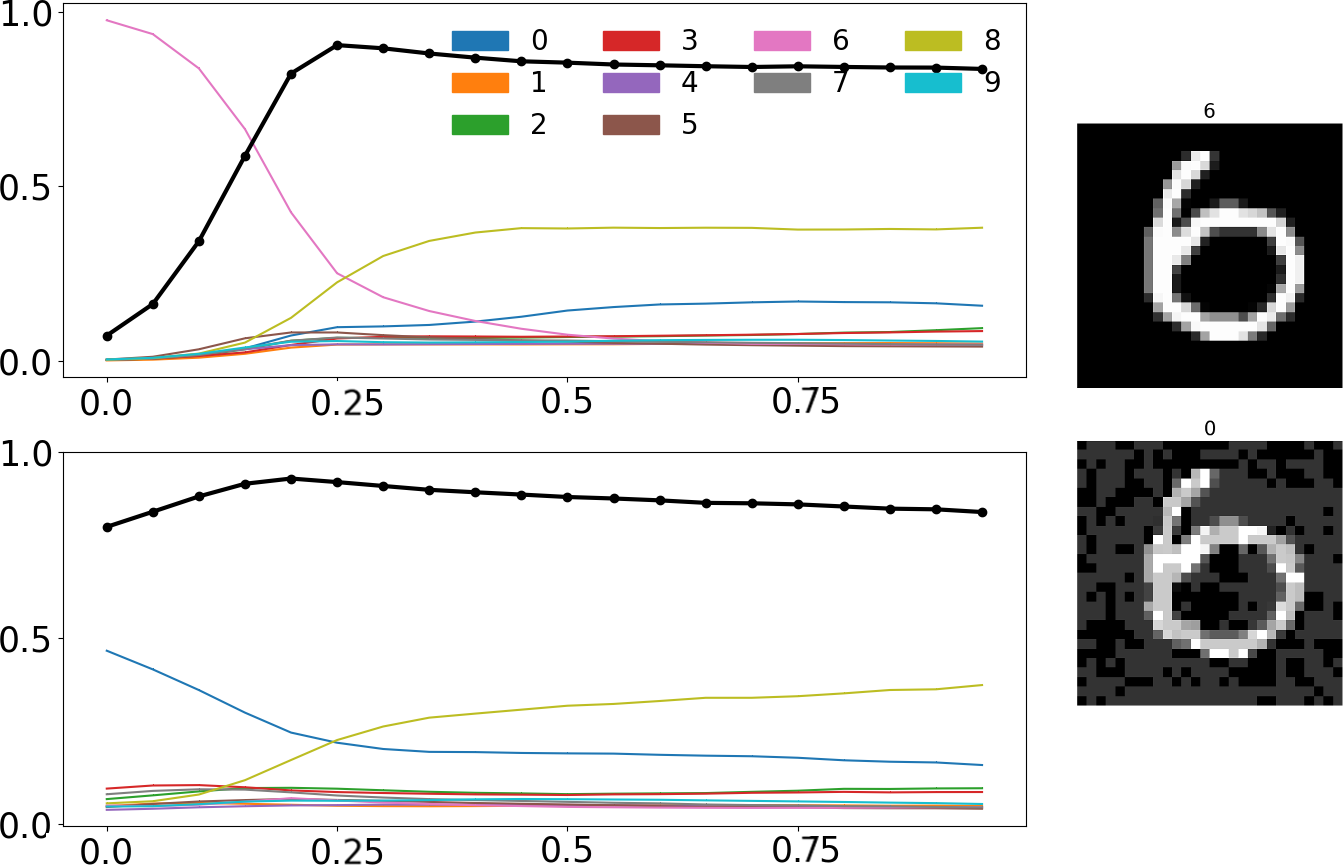}
\end{minipage}\hspace{0.1cm}
\begin{minipage}{.49\linewidth}%
\includegraphics[width=1.07\linewidth]{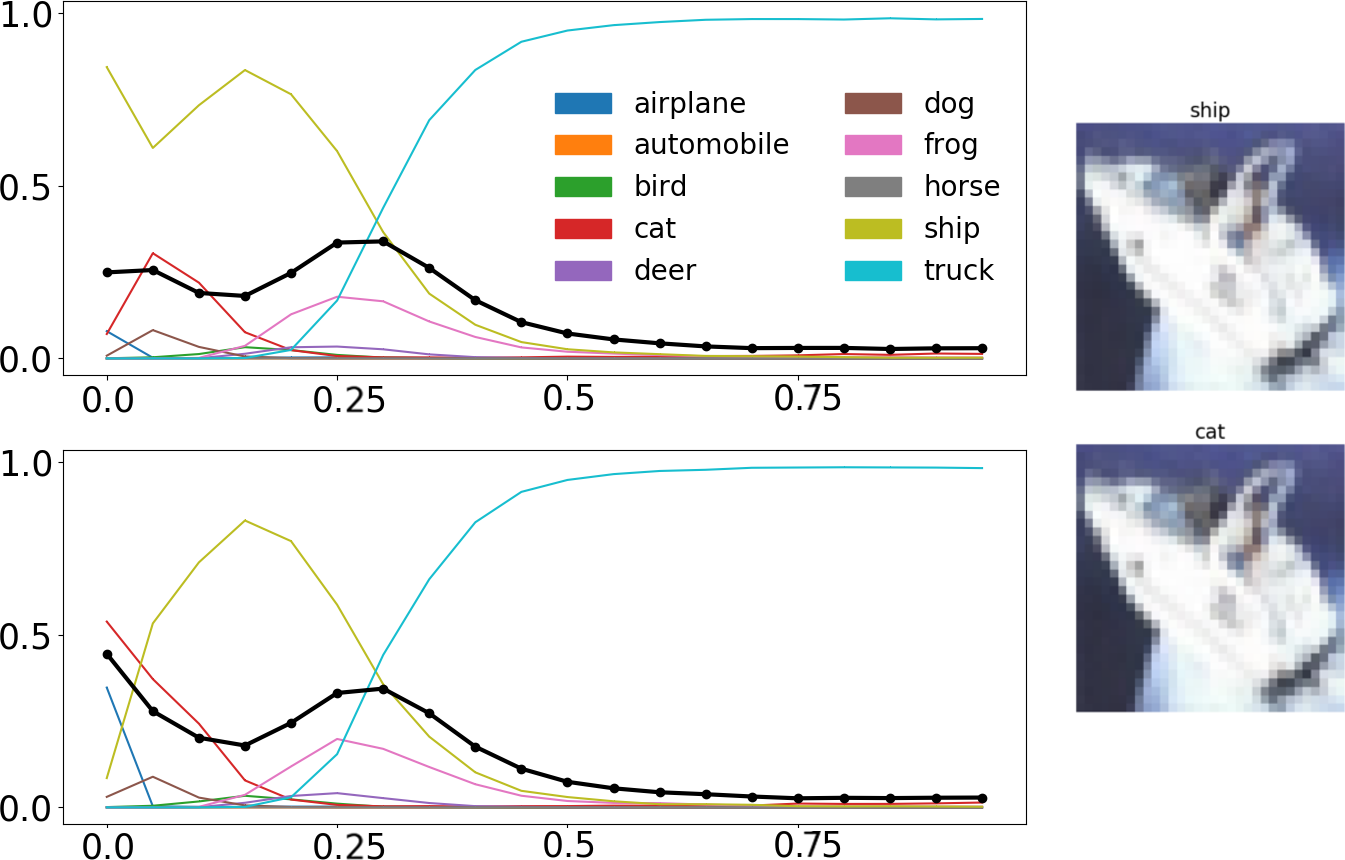}
\end{minipage}\\
\begin{minipage}{.49\linewidth}%
\hspace{2.2cm} radius $r$
\end{minipage}
\begin{minipage}{.49\linewidth}%
\hspace{2.99cm} radius $r$
\end{minipage}\\
\begin{minipage}{.49\linewidth}%
\centering
(a)
\end{minipage}%
\begin{minipage}{.49\linewidth}%
\centering(b)
\end{minipage}
\vspace{-0.3cm}
\caption{\zs values (dotted black line) on $y$-axes, with a varying radius $r\in[0,1]$ on $x$-axes. Mean confidence scores for each class in the explored zone are also reported (colored lines). (a) MNIST dataset: Top Figure: a 6 digit classified by the MLP model. Bottom Figure: a adversarial example classified as a 0. (b) CIFAR-10 dataset: Top Figure: a ship classified by RESNETv2. Bottom Figure: an adversarial example classified as a cat.}
\label{fig:wrt_radius}%
\end{figure*}

\subsection{Uncertainty around a given input}
\label{ss:uncertainty}

A first \zs use case is to assess at which distance and in which proportions are other classes surrounding an interesting input $X$.

Figure \ref{fig:wrt_radius} presents the \zs values with respect various radii $r$, around two input examples.
A radius value $r=0$ is equivalent to the simple output vector corresponding to the inference of $X$, i.e., without actual sampling in $\mathbf{B}$. Our algorithm starts with a $r=0$, up to $1$\footnote{Note that  for $X\in[0,1]^d$ and for $r\geq1$,  $\mathbf{B}_{\boldsymbol{\infty} }(X,r)\cap[0,1]^d = [0,1]^d$, then the space of normalized data is totally covered by the sampling process.
}.
For top-experiment, the confidence value for $r=0$ is high for digit 6 (around 0.9), which is reflected by \zs value to be low (around 0.2). When $r$ increases, the confidence values of other classes are increasing progressively, which results for \zs to increase sharply; at radii of $r=[0.25,1]$ uncertainty in such a ball is critical. 

Adversarial examples were proposed by \cite{Goodfellow:2015}, as attacks against a model: a small perturbation of an input example causes a misclassification by that model. Some of those attacks are close to invisible to humans \cite{DBLP:journals/corr/KurakinGB16}; this makes the use of an index particularly interesting for detection.
The bottom experiment on Figure \ref{fig:wrt_radius} (a) inspects an adversarial example created from the previous digit (we used the \textit{fast gradient method} from \cite{Goodfellow:2015}, with a $\epsilon=0.2$). We remark that at $r=0$, \zs is high, and remains as such with increasing $r$. 
We perform the same experiment on one image from the CIFAR-10 dataset (a ship), and on an adversarial example (classified as a cat, with $\epsilon=0.008$) on Bottom Figure \ref{fig:wrt_radius} (b): the \zs value is twice higher for the adversarial example than for the standard input. 
This exhibits that \zs consistently captures the uncertainty related to the close boundary location for the adversarial examples, and then may allow to identify certain inputs as a problematic one at inference time.

This experiment brings two other observations:  \textit(i) the study of the evolution of \zs values with growing radii is of interest to characterize surrounding or problematic zones (see sharp variations on Figure \ref{fig:wrt_radius}). \textit(ii) for a $r$ covering the whole input space, we obtain an estimate of the surface occupied by each class in the model (e.g., class 8 dominates on MNIST, while class 'truck' is covering most of classification space for the CIFAR-10 case).

\subsection{The surrounding of adversarial examples}
\label{ss:adv}

\begin{figure}[!h]
\centering
\includegraphics[width=1\linewidth]{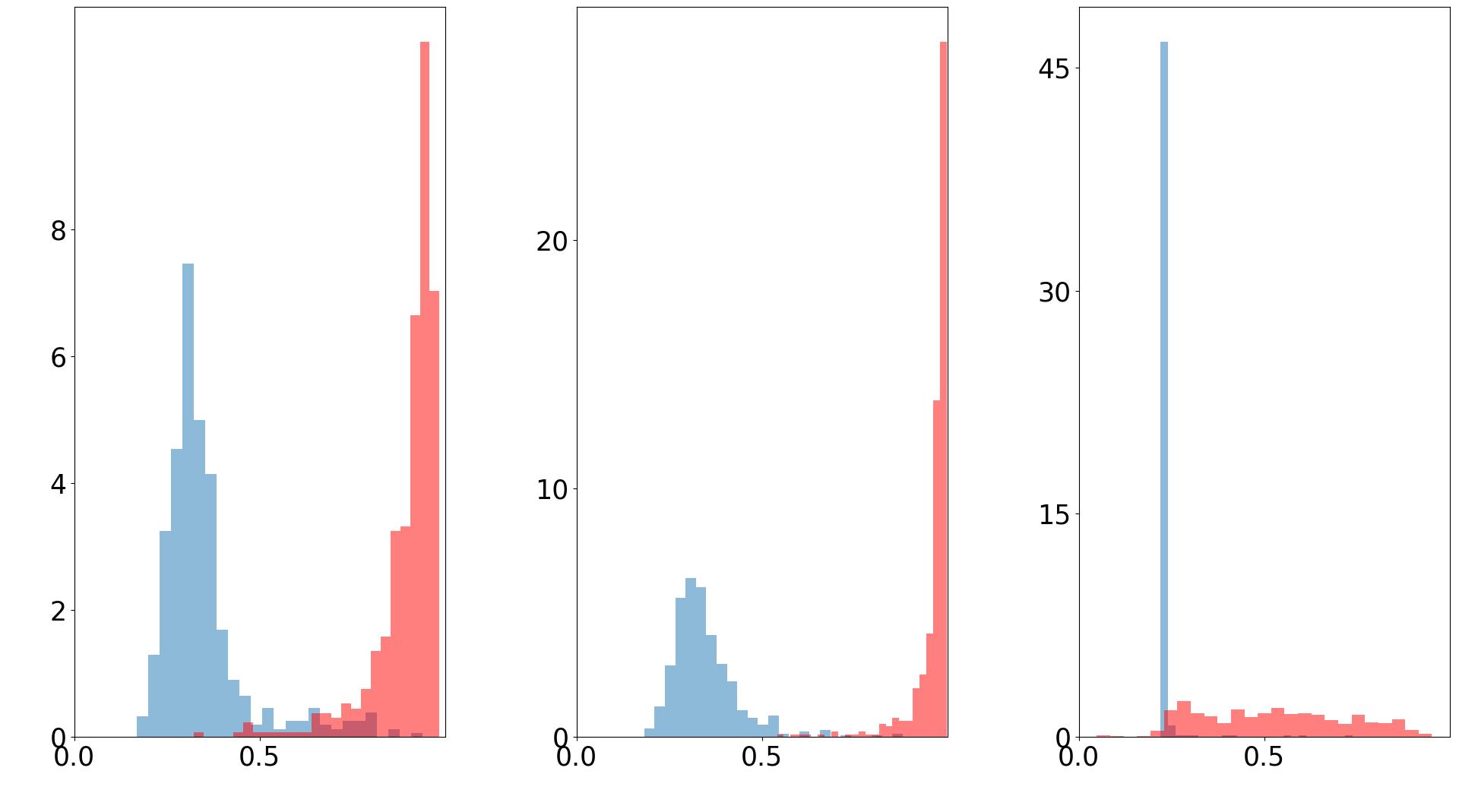}
\caption{Relative distributions of \zs values ($x$-axes) around 500 random test set examples (blue) and around the 500 adversarial examples generated from those test set examples (red), for the three networks (MLP, CNN and IRNN, from left to right, respectively).} 
\label{fig:fgm}
\end{figure}

We now question the relevance of \zs in the task of discriminating adversarial examples from standard (i.e., non-adversarial) inputs.
We add two models for the experiment, also from Keras \cite{keras}: a convolutional neural network (CNN), and a recurrent neural network (IRNN), and focus on the MNIST dataset.
First, the distribution of \zs values (estimated in $\mathbf{B}$ with $r=0.025$) around 500 inputs taken from the test set, are plotted in blue. Those distributions are represented for the three neural networks.
The experiment then creates 500 adversarial examples per model with the fast gradient method, starting from the previous 500 inputs (red distribution). 
 Figure \ref{fig:fgm} plots those empirical distributions. 
 
 We can observe significant difference in those distributions, for each model. The mean of those empirical distributions are reported in Table \ref{tab:means} (rows in grey). Consistently with the experiment in subsection \ref{ss:uncertainty}, we observe higher values of \zs around adversarial examples than around standard inputs. 

In order to formally assess this difference in the two distributions, we performed the statistical \textit{Kolmogorov-Smirnov 2-sample test} \cite{mann1947}; it is e.g., available under R. It tests the null hypothesis that two samples (here of both adversarial and standard inputs \zs values) were drawn from the same continuous distribution. In any of the three case (MLP, CNN and IRNN models), the test rejects the null hypothesis with p-values lower than $10^{-3}$. The p-values of this test correspond to the probabilities of observing the values under the null hypothesis. 
This confirms the distinguishability of the two distributions, and then the interest of \zs. 

\subsection{Disagreement on corner case inputs}
\label{ss:disagreement}
Arguably interesting inputs are the ones that make several deep neural network models disagree, as depicted in the case of self-driving cars \cite{deepxplore}. Indeed, they constitute corner cases, that have to be dealt with for most critical applications.
While this problem may be masked using voting techniques by several independent classifiers \cite{Kuncheva2003}, we are interested in studying the \zs value of those particular inputs. 
Given two models, those inputs are extracted by comparing the predictions of the two models on a given dataset.
In the MNIST dataset, we identify a total of $182$ disagreements (i.e., corner cases) for the three models.
 
\begin{figure}[!h]
\centering
\includegraphics[width=1\linewidth]{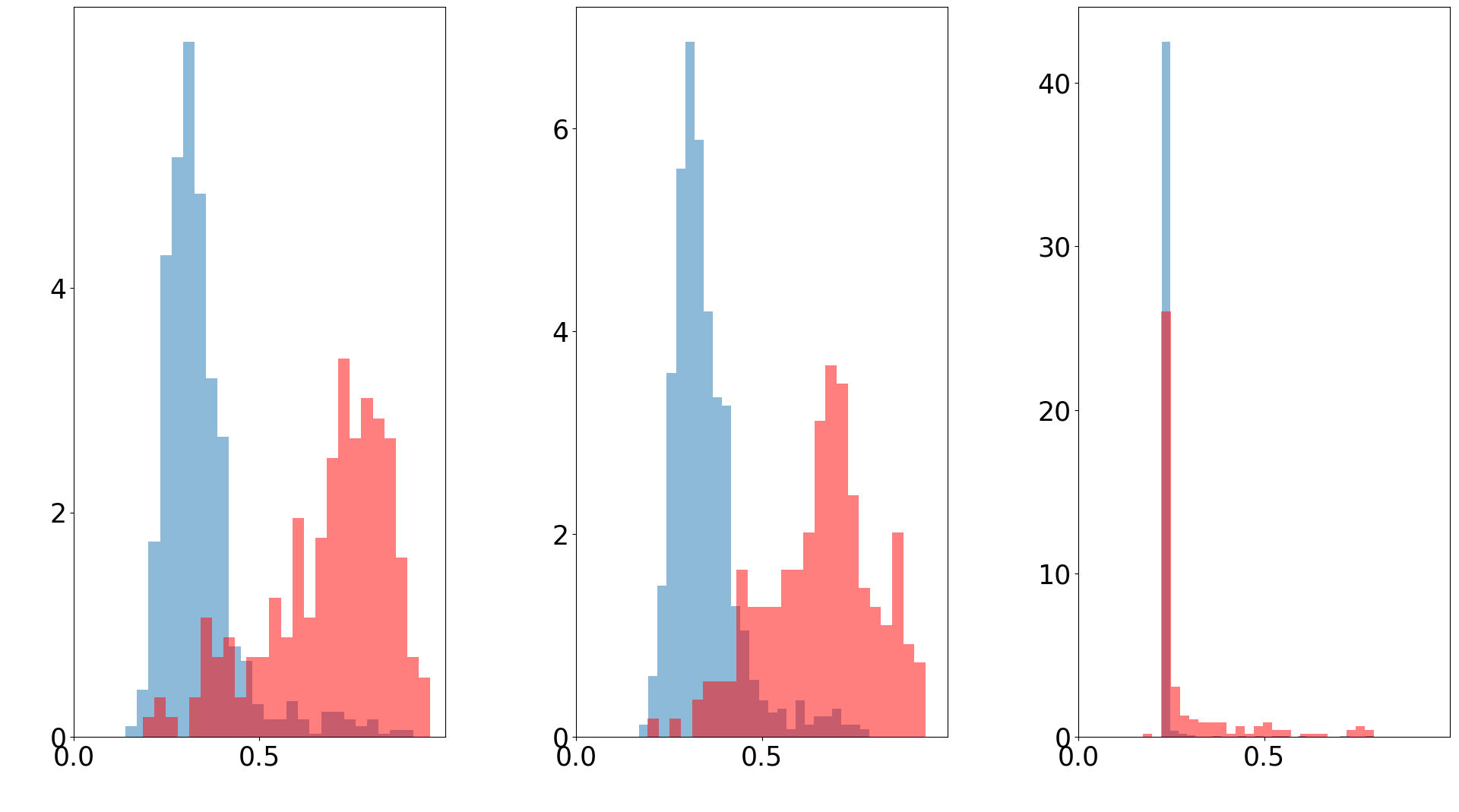}
\caption{Relative distributions of \zs around $1,000$ random test set inputs (blue) and the 182 corner case inputs (red).}
\label{fig:conflict}
\end{figure}
We plot on Figure \ref{fig:conflict} the two following empirical distributions. First, in blue, the distribution of $1,000$  \zs values around test set inputs; second, in red, the distribution of the 182 corner case inputs. The \zs index is also estimated with $r=0.025$. 

As for previous experiment, we observe a significant difference in the distributions (means are reported on Table \ref{tab:means}, rows in white), yet with a larger overlap. The Kolmogorov-Smirnov 2-sample test rejects in each case the null hypothesis with p-values lower than $10^{-3}$.
This experiment indicates the likelihood to detect those critical inputs due to their higher value, while running inference on a single network.

\subsection{Effects of watermarking on model boundaries}
\label{ss:watermarking}

As a final illustration, we consider the recent work \cite{watermarking} consisting in \textit{watermarking} a model. This operation is performed as follows: first, a watermark key consisting of inputs is created. Half of the key inputs are actually adversarial examples. Second, the watermarking step consists in finetuning the model by retraining it over the key inputs, in order to re-integrate the adversarial examples in their original class. That last step is by definition moving the model decision boundaries, which constitutes the watermark that the model owner has sought to introduce.

All three models are watermarked with keys of size $100$. Each input of those three keys is used as input $X$ for \zs, for $100$ runs of Algorithm \ref{algo} with $k=1,000$ samples each. Two distributions of \zs values are computed: one for the un-marked model, and another for the watermarked one.

If the watermarked model were indistinguishable from the original one, the two distributions would be the same. 
Once again, the Kolmogorov-Smirnov 2-sample test returns p-values lower than $10^{-3}$.
This reflects the fact that boundaries have changed around the key inputs, at least enough to be observed by \zs. As the watermarking process is expected to be as stealthy possible, this highlights that the key is to remain a secret for the model owner only \cite{watermarking}. 

\begin{table}
\centering
\begin{tabular}{|l||ccc|}
\hline
Scenario & MLP & CNN & IRNN \\ 
\hline\hline
\rowcolor{black!20} 500 inputs from test set & 0.3499 & 0.3462 & 0.2399  \\ \hline
\rowcolor{black!20} 500 adversarial examples  & 0.8986 & 0.9613 & 0.5254   \\ \hline
1,000 inputs from test set  & 0.3709 & 0.3457 & 0.2403   \\ \hline
182 corner case inputs  & 0.6857 & 0.6564 & 0.2989   \\ \hline
\end{tabular}
\caption{Mean of the empirical \zs distributions, for the three neural networks. The grey rows refer to experiments in subsection \ref{ss:adv}, and the white ones to \ref{ss:disagreement}.}
\label{tab:means}
\vspace{-0.9cm}
\end{table}

\section{Related works}
\label{s:related}

While some metrics such as distance-from-boundaries \cite{941248} have been developed for support vector machines for instance, the difficulty to reason about the classifier decision boundaries of neural networks have motivated research works since decades.
Lee et al. \cite{boundary-extraction} proposed the extraction of features from decision boundaries, followed by \cite{554193} that targets the same operation without assuming underlying probability distribution functions of the data.
An interesting library for visual inspection of boundaries, from low dimensional embedding is available online \cite{plotting}.
Fawzi et al. propose to analyze some geometric properties of deep neural network classifiers \cite{DBLP:journals/corr/FawziMFS17},  including the curvature of those boundaries.
Van den Berg \cite{insights} studies the decision region formation in the specific case of feedforward neural networks with sigmoidal nonlinearities.

Both Pei et al. \cite{deepxplore} and Goodfellow et al. \cite{Goodfellow:2015} propose to generate inputs that may induce problems in the tested deep neural networks. Those are proposing attacks, not indexes.

Cross entropy is one of the most basic measure for the training of neural networks. While it evaluates the divergence of the dataset distribution from the model distribution, it is not applied to inspect the confidence in the inference of specific zones in the input space.  
This paper addresses the relation of the surrounding of a given input with respect to the presence of decision boundaries of other classes; this relation is captured by the \zs index, leveraging the Shannon entropy.

\section{Conclusion}
\label{s:conclusion}
We introduced a novel index, \zs, for inspecting the surrounding of given inputs with regards to the boundary entropy measured in the zones of interest. We have presented a Monte Carlo estimation of that index, and have shown its applicability on a base case, on a recent technique for model watermarking, and on two concerns regarding the adoption of current deep neural networks (adversarial examples and corner case inputs).
As this index is intended to be generic, we expect other applications to leverage it; future works for instance include the study of the link of \zs values with regards to other identified issues such as new \textit{trojaning} attacks on neural networks \cite{Trojannn}.

\bibliographystyle{IEEEtran}
\bibliography{mddl}

\begin{thebibliography}{10}
\providecommand{\url}[1]{#1}
\csname url@samestyle\endcsname
\providecommand{\newblock}{\relax}
\providecommand{\bibinfo}[2]{#2}
\providecommand{\BIBentrySTDinterwordspacing}{\spaceskip=0pt\relax}
\providecommand{\BIBentryALTinterwordstretchfactor}{4}
\providecommand{\BIBentryALTinterwordspacing}{\spaceskip=\fontdimen2\font plus
\BIBentryALTinterwordstretchfactor\fontdimen3\font minus
  \fontdimen4\font\relax}
\providecommand{\BIBforeignlanguage}[2]{{%
\expandafter\ifx\csname l@#1\endcsname\relax
\typeout{** WARNING: IEEEtran.bst: No hyphenation pattern has been}%
\typeout{** loaded for the language `#1'. Using the pattern for}%
\typeout{** the default language instead.}%
\else
\language=\csname l@#1\endcsname
\fi
#2}}
\providecommand{\BIBdecl}{\relax}
\BIBdecl

\bibitem{995823}
Z.~Wang and A.~C. Bovik, ``A universal image quality index,'' \emph{IEEE Signal
  Processing Letters}, vol.~9, no.~3, pp. 81--84, March 2002.

\bibitem{941248}
G.~Guo, H.-J. Zhang, and S.~Z. Li, ``Distance-from-boundary as a metric for
  texture image retrieval,'' in \emph{IEEE International Conference on
  Acoustics, Speech, and Signal Processing}, vol.~3, 2001, pp. 1629--1632
  vol.3.

\bibitem{DBLP:journals/corr/FawziMFS17}
A.~Fawzi, S.~Moosavi{-}Dezfooli, P.~Frossard, and S.~Soatto, ``Classification
  regions of deep neural networks,'' \emph{CoRR}, vol. abs/1705.09552, 2017.

\bibitem{deepxplore}
K.~Pei, Y.~Cao, J.~Yang, and S.~Jana, ``Deepxplore: Automated whitebox testing
  of deep learning systems,'' in \emph{SOSP}, 2017.

\bibitem{Goodfellow:2015}
I.~J. Goodfellow, J.~Shlens, and C.~Szegedy, ``Explaining and harnessing
  adversarial examples,'' in \emph{ICLR}, 2015.

\bibitem{watermarking}
E.~{Le Merrer}, P.~Perez, and G.~Tr\'edan, ``Adversarial frontier stitching for
  remote neural network watermarking,'' \emph{CoRR}, vol. abs/1711.01894, 2017.

\bibitem{plotting}
``Plotting high-dimensional decision boundaries,''
  \url{https://github.com/tmadl/highdimensional-decision-boundary-plot},
  accessed: 2018-07-01.

\bibitem{source}
``{zoNNscan}: a boundary-entropy metric for zone inspection of neural models,''
  \url{https://github.com/technicolor-research/zoNNscan}, uploaded: 2018-08-16.

\bibitem{insights}
E.~van~den Berg, ``Some insights into the geometry and training of neural
  networks,'' \emph{CoRR}, vol. abs/1605.00329, 2016.

\bibitem{554193}
C.~Lee and D.~A. Landgrebe, ``Decision boundary feature extraction for neural
  networks,'' \emph{IEEE Transactions on Neural Networks}, vol.~8, no.~1, pp.
  75--83, Jan 1997.

\bibitem{dick_kuo_sloan_2013}
J.~Dick, F.~Y. Kuo, and I.~H. Sloan, ``High-dimensional integration: The
  quasi-monte carlo way,'' \emph{Acta Numerica}, vol.~22, pp. 133--288, 2013.

\bibitem{keras}
``Keras: Deep learning for humans,''
  \url{https://github.com/keras-team/keras/tree/master/examples}, accessed:
  2018-01-01.

\bibitem{DBLP:journals/corr/KurakinGB16}
A.~Kurakin, I.~J. Goodfellow, and S.~Bengio, ``Adversarial examples in the
  physical world,'' \emph{CoRR}, vol. abs/1607.02533, 2016.

\bibitem{mann1947}
H.~B. Mann and D.~R. Whitney, ``On a test of whether one of two random
  variables is stochastically larger than the other,'' \emph{Ann. Math.
  Statist.}, vol.~18, no.~1, pp. 50--60, 03 1947.

\bibitem{Kuncheva2003}
L.~Kuncheva, C.~Whitaker, C.~Shipp, and R.~Duin, ``Limits on the majority vote
  accuracy in classifier fusion,'' \emph{Pattern Analysis {\&} Applications},
  vol.~6, no.~1, pp. 22--31, Apr 2003.

\bibitem{boundary-extraction}
C.~Lee and D.~A. Landgrebe, ``Feature extraction based on decision
  boundaries,'' \emph{IEEE Transactions on Pattern Analysis and Machine
  Intelligence}, vol.~15, no.~4, pp. 388--400, Apr 1993.

\bibitem{Trojannn}
Y.~Liu, S.~Ma, Y.~Aafer, W.-C. Lee, J.~Zhai, W.~Wang, and X.~Zhang, ``Trojaning
  attack on neural networks,'' in \emph{NDSS}, 2018.

\end{thebibliography}

\end{document}